\documentclass{article}
\usepackage[main,final]{neurips_2026}
\usepackage[utf8]{inputenc}
\usepackage[T1]{fontenc}

\usepackage{amsmath}
\usepackage{amssymb}
\usepackage{amsthm}
\usepackage{mathtools}
\usepackage{bm}
\usepackage{nicefrac}

\usepackage{microtype}

\usepackage[table]{xcolor}
\usepackage{graphicx}
\usepackage{subcaption}
\usepackage{booktabs}
\usepackage{tabularx}
\usepackage{multirow}
\usepackage{makecell}

\usepackage{algorithm}
\usepackage{algpseudocode}

\usepackage{wrapfig}
\usepackage{placeins}
\usepackage{float}

\usepackage{tikz}
\usepackage{pgfplots}
\pgfplotsset{compat=1.17}
\usetikzlibrary{positioning,fit,calc,backgrounds,patterns,matrix}

\usepackage{dirtytalk}
\usepackage{comment}
\usepackage{enumerate}
\definecolor{myRed}{HTML}{D81B60}
\definecolor{myBlue}{HTML}{1E88E5}
\definecolor{myYellow}{HTML}{FFC107}
\definecolor{myGreen}{HTML}{007B67}

\usepackage{tcolorbox}

\newtcbox{\myGreenHighlight}[1][]{nobeforeafter, tcbox raise base,
  boxrule=0pt, top=0pt, bottom=0pt, left=0mm, right=0mm,
  arc=0pt, auto outer arc, boxsep=0pt, leftright skip=0pt,
  colback=myGreen!20, colframe=myGreen!20, #1}

\usepackage{url}
\usepackage{hyperref}
\hypersetup{
    colorlinks=true,
    linkcolor=blue,
    filecolor=magenta,
    urlcolor=cyan,
    citecolor=blue
}

\usepackage[capitalize,noabbrev]{cleveref}

\crefname{proposition}{Proposition}{Propositions}
\Crefname{proposition}{Proposition}{Propositions}
\theoremstyle{plain}
\newtheorem{theorem}{Theorem}[section]
\newtheorem{proposition}[theorem]{Proposition}

\theoremstyle{definition}

\theoremstyle{remark}

\title{Adaptive Entropy-Sparing for Efficient Reasoning}

\author{%
  Cong Jiang \textsuperscript{1,2}\quad
  Xiaofeng Zhang \textsuperscript{1}\quad
  Tom Ko \textsuperscript{2}\quad
  Zheng Zhang \textsuperscript{1,2}%
  \thanks{Corresponding author.}\\
  \textsuperscript{1}Harbin Institute of Technology, Shenzhen\\
  \textsuperscript{2}Shenzhen Loop Area Institute\\
  \texttt{jerryjiang.jc@gmail.com},
  \texttt{darrenzz219@gmail.com}
}

\begin{document}

\maketitle

\begin{abstract}
Large Reasoning Models (LRMs) produce explicit chains of thought to improve transparency, but they frequently overthink, generating long yet inefficient reasoning chains that inflate computational costs and risk introducing errors. 
While numerous reinforcement learning methods struggle to penalize verbosity to achieve conciseness, they typically apply the penalty uniformly across lengthy sequences, which leads to \textit{credit confounding}, that is, the failure to distinguish essential reasoning steps from superfluous ones. In this paper, we study a measurable proxy for this confounding: high-entropy tokens that often coincide with surface transition points where reasoning re-enters verification.
Based on this insight, we propose Adaptive Entropy-Sparing (AES), which selectively spares high-entropy tokens that occur before an efficient in-group reference, while penalizing overlength high-entropy suffix tokens associated with superfluous branching.
Our AES thereby navigates the suppression-exploration trade-off, reducing overthinking while preserving average reasoning accuracy. We also present an entropy-length model that motivates why local uncertainty can serve as a proxy for future reasoning cost.
Across eight reasoning benchmarks, AES reduces reasoning length by up to 52.9\% while increasing mean accuracy by up to 3.8\% over the base models, consistently outperforming existing efficiency-oriented methods.
Our code is available at https://github.com/MobiusDai/AES.
\end{abstract}

\section{Introduction}
Large reasoning models (LRMs) have recently demonstrated remarkable problem-solving capabilities, particularly in mathematical reasoning~\citep{reasoning_math}, code generation~\citep{reasoning_code}, and agentic tasks~\citep{reasoning_agent}. The reasoning ability of large language models (LLMs) can be traced back to chain-of-thought (CoT) prompting~\citep{cot}, which enables models to solve problems step-by-step through simple instructions without additional fine-tuning. While such prompt-based reasoning represents an early form of deliberative thinking, recent advances have given rise to slow-thinking models~\citep{slowthinking}, such as OpenAI-o1~\citep{openai_o1}, DeepSeek-R1~\citep{deepseek}, and QWQ~\citep{qwq}, which analyze problems through extended reasoning trajectories before producing final answers. Drawing an analogy to dual-process theory in cognitive psychology, these models are often characterized as \textit{System-2} reasoners, in contrast to conventional LLMs that produce direct responses akin to \textit{System-1} processing~\citep{reasoningsystem}. Although LRMs substantially outperform their predecessors, this improvement comes at the cost of significantly increased token consumption. In particular, these models frequently exhibit overthinking~\citep{overthinking}, characterized by redundant verification steps and unnecessary backtracking that inflate computational costs without improving, and sometimes even degrading answer quality.

To address the overthinking problem, various methods have been proposed to reduce output length while maintaining reasoning quality~\citep{efficient_survey}. Existing techniques can be broadly categorized into three research lines: efficiency-oriented post-training~\citep{l1, thinkprune, lcr1}, latent reasoning~\citep{coconut,latent_thought,lrt}, and model collaboration~\citep{sd_efficientreasoning, entropy_sd_reasoning}. Among these, efficiency-oriented post-training is the most widely adopted, as it preserves the original model architecture and does not require auxiliary models. These methods typically augment outcome-based rewards with cost-aware terms~\citep{o1pruner, laser}, incentivizing the model to produce more concise responses. 

Despite their efficacy, these approaches often suffer from performance degradation~\citep{efficient_reason} compared to the original models, particularly on challenging tasks. Cost-aware penalties uniformly penalize all tokens in a trajectory, conflating steps that are essential for correct reasoning with those that merely introduce superfluous elaboration. We refer to this limitation as \textit{credit confounding}. Consequently, the model is discouraged from engaging in deliberate, multi-step reasoning even when such exploration is essential for correctness, leading to a suboptimal suppression-exploration trade-off.

In this work, we begin by analyzing model-generated reasoning trajectories through token-level conditional entropy. We observe that transition words and phrases (\textit{e.g.,} `but,' `wait,' `let us verify') frequently align with high-entropy spikes, indicating elevated local uncertainty. These high-entropy tokens serve as transition triggers that may either facilitate essential reasoning or induce superfluous exploration. When the model hesitates among multiple candidate directions, these tokens often initiate additional verification steps rather than concluding the reasoning process~\citep{s1}. To further investigate this phenomenon, we conduct controlled experiments by manually truncating reasoning trajectories at different transition points. These results reveal that later transition steps contribute minimally to final answer quality, and a substantial portion of high-entropy transitions induce redundant branching rather than productive exploration. These observations yield a key insight: \textit{overthinking is often associated with the model's insufficient confidence to terminate reasoning after reaching a plausible conclusion, where high-entropy tokens guide the model toward unnecessary subsequent verification steps}.

Building on this insight, we propose Adaptive Entropy-Sparing (AES), a reinforcement learning based post-training method that applies selective advantage reduction to overlength high-entropy suffix tokens in correct trajectories. To operationalize this idea, we sample multiple responses for each prompt and treat the shortest correct response as a reference for efficient reasoning, which allows us to estimate the degree of overthinking in longer correct responses and apply penalties selectively. Extensive experiments demonstrate that AES substantially reduces response length while improving accuracy across both mathematical reasoning and out-of-domain benchmarks.

\section{Empirical Analysis of Overthinking}
\label{observation}
To investigate patterns associated with the overthinking phenomenon in LRMs, we analyze reasoning trajectories from an information-theoretic perspective. We examine whether overthinking is associated with identifiable patterns in the model's token-level uncertainty during generation. To do so, we first inspect the entropy distribution across tokens within reasoning trajectories and find that specific transitional markers frequently exhibit elevated entropy. We then conduct truncation experiments with generated reasoning trajectories to determine whether extended reasoning steps contribute meaningfully to final answer quality.

\subsection{Entropy Distribution Across Reasoning Trajectories}

\textbf{Settings.} Let $\pi_\theta$ denote a slow-thinking language model. In our implementation, we employ DeepSeek-R1-Distill-Qwen-7B~\citep{deepseek} and evaluate on the MATH-500~\citep{Math500} benchmark. Given a prompt $x$, the generation process is formalized as $[R, y] \sim \pi_\theta(\cdot \mid x)$, where $R = [r_1, r_2, \dots, r_n]$ represents the reasoning trajectory, $y$ denotes the summarized answer, and $[\cdot, \cdot]$ indicates concatenation. Under greedy decoding, the output is deterministically obtained as $y \sim \pi_\theta(\cdot \mid [x, R])$.
Our analysis focuses on the token-level uncertainty within $R$. When generating the $i$-th token $r_i$, the conditional distribution is given by $\pi(r_i) = \pi_\theta(r_i \mid x, r_1, \dots, r_{i-1})$. We quantify the local uncertainty at position $i$ using the conditional entropy:
\begin{equation}
H_i = -\sum_{v \in \mathcal{V}} \pi(v) \log \pi(v),
\end{equation}
where $\mathcal{V}$ denotes the vocabulary. To identify high-entropy segments, which we treat as candidate high-uncertainty regions, we employ a relative thresholding strategy. A token is classified as high-entropy if its entropy value falls within the top 5\% of the trajectory. To ensure robustness and filter out noise, we retain a segment only if the majority of tokens within a five-token sliding window satisfy this criterion.

\begin{wrapfigure}{r}{0.48\textwidth}
  \vspace{-0.3em}
  \centering
  \includegraphics[width=\linewidth]{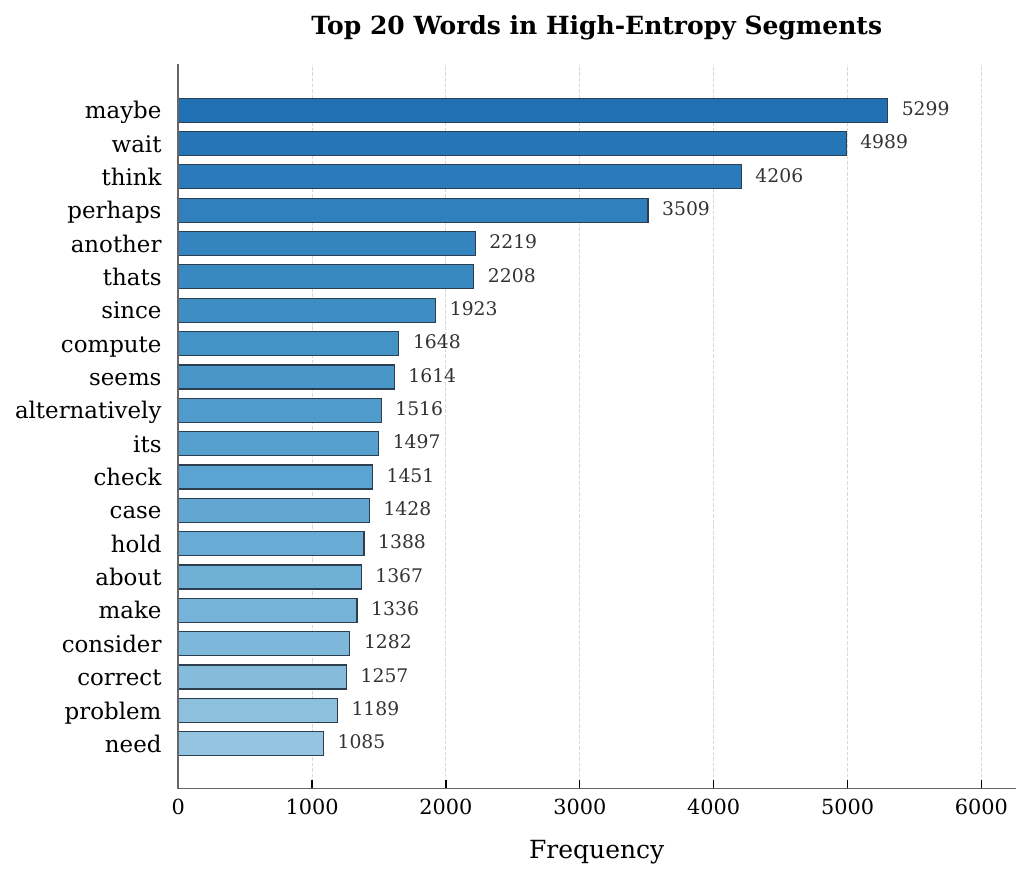}
  \caption{
    Top-20 most frequent tokens in high-entropy reasoning segments.
  }
  \label{fig:entropy_visualization}
  \vspace{-0.3em}
\end{wrapfigure}

\textbf{Observations.} As illustrated in \cref{fig:entropy_visualization}, the vocabulary distribution within high-entropy segments is highly non-uniform, predominantly featuring what we term \textit{transition markers} rather than domain-specific terminology. The most frequent high-entropy tokens serve as distinct markers of uncertainty and redirection. We observe a strong prevalence of uncertainty markers such as ``maybe'', ``perhaps'', and ``seems'', which are consistent with candidate branching points where the model deliberates between multiple reasoning paths. Furthermore, the high frequency of self-correction markers such as ``wait'', ``check'', and ``hold'' suggests that elevated entropy correlates with moments of retrospective verification. These patterns indicate that overthinking is characterized by recursive cycles of hesitation and self-correction, where substantial probability mass is allocated to re-evaluating prior reasoning steps rather than advancing toward the solution.

\subsection{Impact of Trajectory Truncation on Answer Quality}

The preceding analysis suggests that high-entropy segments correspond to moments of model uncertainty and potential branching. A natural follow-up question arises: \textit{do the reasoning steps that follow these transition points contribute meaningfully to the final answer?} To address this question, we design a truncation experiment.

\begin{figure}[ht]
  \begin{center}
    \centerline{\includegraphics[width=0.6\columnwidth]{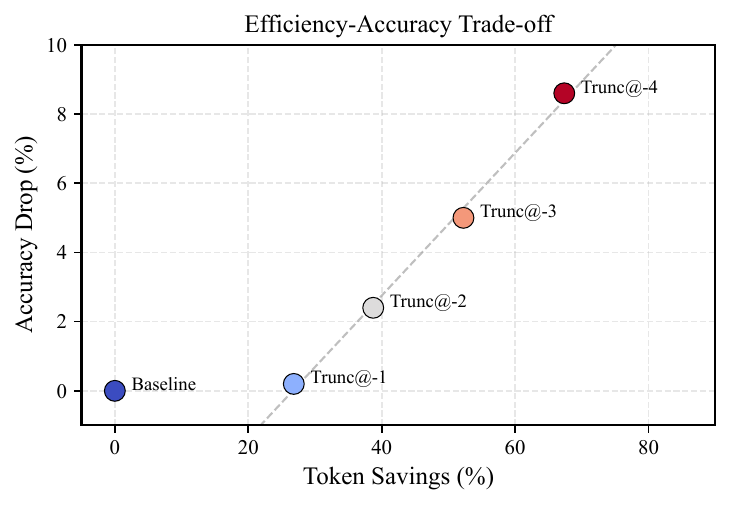}}
    \caption{
      Impact of reasoning trajectory truncation on model performance. The plot compares the percentage of token savings against the accuracy drop relative to the baseline.
    }
    \label{efficiency_tradeoff}
  \end{center}
  \vspace{-2em}
\end{figure}

\textbf{Settings.} We identify several high-entropy tokens within a complete reasoning trajectory $R$ using the criteria defined above. These tokens, which we refer to as \textit{transition points}, partition the trajectory into sequential segments. We construct four experimental conditions, denoted as \textit{Trunc@$-k$} for $k \in \{1, 2, 3, 4\}$, where the final $k$ segments are discarded. For each condition, the model generates the final answer $y$ based solely on the truncated prefix. We evaluate performance using two metrics: accuracy drop relative to the full-trajectory baseline and token savings measured as the percentage reduction in reasoning tokens.

\textbf{Observations.} \cref{efficiency_tradeoff} presents the trade-off between computational cost and model performance, providing evidence consistent with some late reasoning segments being redundant. Most notably, the \textit{Trunc@$-1$} condition demonstrates that removing only the final segment yields a substantial 26.8\% reduction in token consumption while incurring a negligible accuracy drop of merely 0.2\% (from 92.8\% to 92.6\%). This result suggests that the model typically resolves the core problem well before concluding its generation, which renders the final reasoning phase computationally redundant in many examples. Furthermore, even under the aggressive truncation of \textit{Trunc@$-4$}, where 67.4\% of reasoning tokens are removed, the model retains the majority of its problem-solving capability, suffering only an 8.6\% decline in accuracy.

\subsection{Summary and Implications}
These findings collectively suggest that LRMs exhibit significant overthinking behavior, as a substantial portion of the generated reasoning, especially in later stages, contributes minimally to answer correctness despite substantial computational cost. Crucially, our entropy analysis suggests that this redundancy is not uniformly distributed but rather concentrated around high-entropy transition points. This observation motivates our proposed approach: \textit{rather than applying uniform length penalties that indiscriminately suppress all tokens, we focus credit reduction on late high-entropy regions that are empirically associated with unnecessary reasoning extensions.}

\section{Methodology}

\begin{figure*}[ht]
  \begin{center}
    \centerline{\includegraphics[width=\columnwidth]{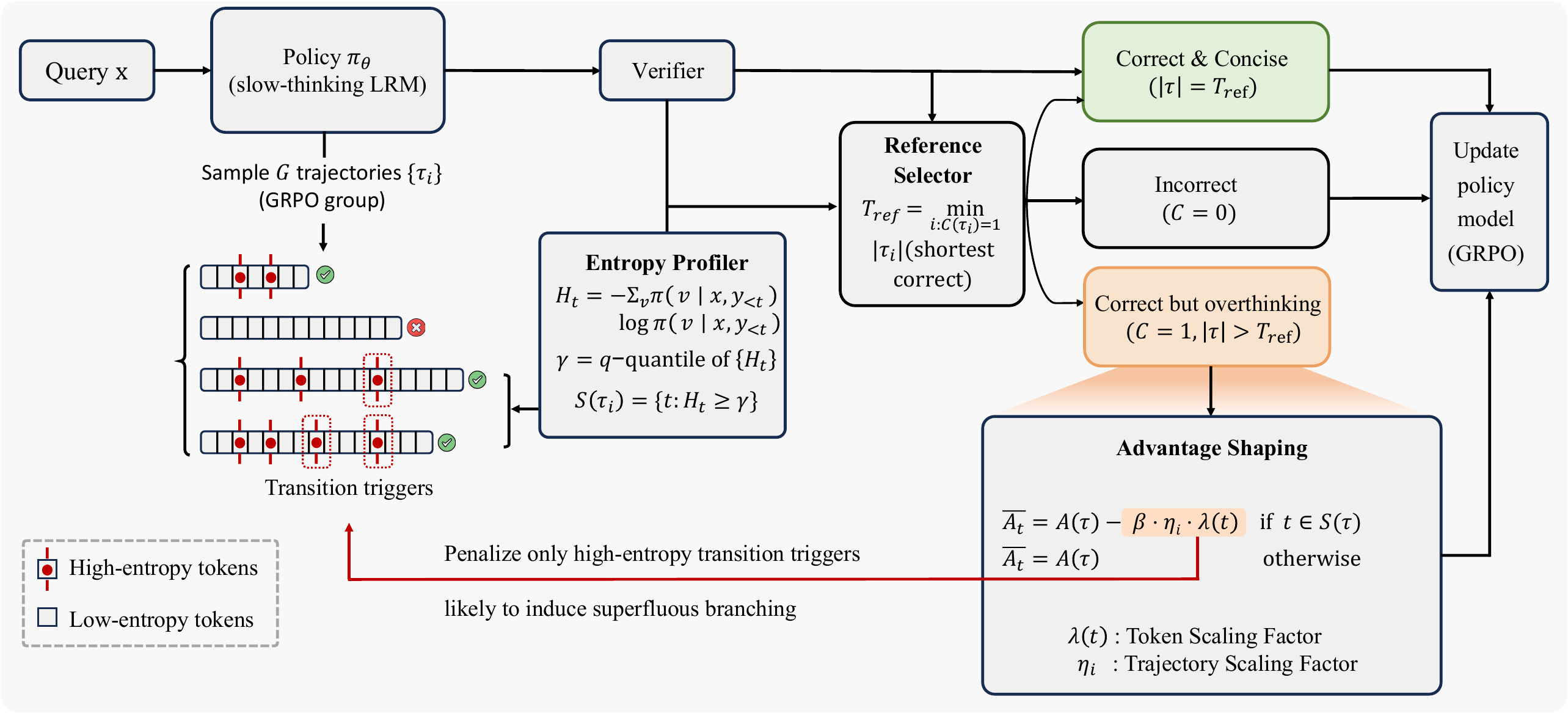}}
    \caption{Overview of Adaptive Entropy-Sparing (AES). AES navigates the suppression-exploration trade-off by selectively penalizing overlength high-entropy suffix tokens associated with superfluous branching, while preserving the original learning signal for concise correct trajectories.}
    \label{pipeline}
  \end{center}
  \vspace{-2em}
\end{figure*}

\subsection{Preliminaries}
\textbf{Reinforcement Learning with Verifiable Rewards.} Reinforcement learning has emerged as an effective paradigm for enhancing the reasoning capabilities of LLMs across diverse domains. In settings where ground-truth answers are available, rule-based outcome rewards provide a simple yet effective training signal. Following prior work, we adopt a reward function defined as: $R(x, y) = +1$ for correct responses, $R(x, y) = -1$ for incorrect responses, where $x$ denotes the input query and $y$ denotes the model response.

The standard RL objective with KL regularization can be formulated as:
\begin{equation}
\max_\theta \mathbb{E}_{x \sim \mathcal{D}} \mathbb{E}_{y \sim \pi_\theta(\cdot|x)} \left[ R(x, y) \right] - \beta_{\mathrm{KL}} D_{\mathrm{KL}}\left[\pi_\theta(\cdot|x) \| \pi_{\mathrm{ref}}(\cdot|x)\right],
\end{equation}
where $\pi_{\mathrm{ref}}$ denotes the reference model and $\beta_{\mathrm{KL}}$ controls the regularization strength.

\textbf{Group Relative Policy Optimization.} We adopt Group Relative Policy Optimization (GRPO)~\citep{reasoning_math} as our base training algorithm. For each query $x$, the behavior policy $\pi_{\theta_{\mathrm{old}}}$ samples a group of $G$ responses $\{y_i\}_{i=1}^G$. The advantage for the $i$-th response is computed by normalizing rewards within the group:
\begin{equation}
A_i = \frac{R_i - \text{mean}(\{R_j\}_{j=1}^G)}{\text{std}(\{R_j\}_{j=1}^G)},
\end{equation}
where $R_i = R(x, y_i)$. The GRPO objective is then formulated as:
\begin{equation}
\mathcal{J}(\theta) = \mathbb{E}_{x \sim \mathcal{D}, y_i \sim \pi_{\theta_{\mathrm{old}}}(\cdot|x)} \bigg[ \frac{1}{G} \sum_{i=1}^G \frac{1}{|y_i|} \sum_{t=1}^{|y_i|}\min\left( \rho_{i,t}(\theta) A_i, \mathrm{clip}(\rho_{i,t}(\theta), 1-\epsilon, 1+\epsilon) A_i \right) \bigg]
\end{equation}
where $\rho_{i,t}(\theta) = \frac{\pi_\theta(y_{i,t} | x, y_{i,<t})}{\pi_{\theta_{\mathrm{old}}}(y_{i,t} | x, y_{i,<t})}$ is the importance sampling ratio and $\epsilon$ is the clipping threshold. We omit the KL term for readability.

\subsection{Limitations of Global Length Penalties}

A common approach to mitigate overthinking is to incorporate a global length penalty into the reward function, penalizing longer responses regardless of content. The modified objective is:
\begin{equation}
\max_\theta \ \mathbb{E}_{\tau \sim \pi_\theta} \left[ R(\tau) - \beta_{\mathrm{len}} \cdot |\tau| \right],
\end{equation}
where $|\tau|$ denotes the total number of tokens in trajectory $\tau$ and $\beta_{\mathrm{len}}$ is the length penalty coefficient.

\textbf{Credit Confounding.} While conceptually simple, lengthy responses often contain substantial portions of valid intermediate reasoning steps that are essential for reaching the correct answer. The global length penalty uniformly penalizes all tokens in the trajectory, which fails to distinguish between tokens that contribute to sound reasoning and those that belong to unnecessary elaboration. 

\subsection{Adaptive Entropy-Sparing}

To address credit confounding, we propose Adaptive Entropy-Sparing (AES), which preserves high-entropy tokens in concise or pre-reference regions while penalizing overlength high-entropy suffix tokens. AES is not a classifier of every transition token. Instead, it implements a conservative selective-penalization rule: extra penalties are applied only when a token is high-entropy, appears in a correct trajectory that is longer than an in-group efficient reference, and lies in the overlength suffix beyond that reference. 
This conservative rule improves the suppression-exploration trade-off by reducing overthinking while preserving useful exploration in concise correct trajectories. An overview of our method is illustrated in \cref{pipeline}.

Our method operates in three stages: (1) identifying high-entropy tokens within each trajectory, (2) classifying trajectories based on correctness and length, and (3) applying differentiated reward shaping according to trajectory type. Here, a trajectory denotes the full generated completion, including the reasoning trace and final answer.

\subsubsection{High-Entropy Token Identification}

For each sampled trajectory $\tau = (y_1, \ldots, y_T)$, we compute the token-level conditional entropy at each generation step from the rollout logits produced by the behavior policy that sampled the trajectory:
\begin{equation}
H_t = -\sum_{v \in \mathcal{V}} \pi_{\theta_{\mathrm{old}}}(v | x, y_{<t}) \log \pi_{\theta_{\mathrm{old}}}(v | x, y_{<t}),
\end{equation}
where $\mathcal{V}$ denotes the vocabulary. This uses information already available during rollout collection and keeps the shaped advantages fixed during the subsequent policy update. We then identify the set of high-entropy tokens as:
\begin{equation}
\mathcal{S}(\tau) = \{ t : H_t \geq \gamma \},
\end{equation}
where $\gamma$ is the entropy threshold. 

\textbf{Threshold Selection.} The choice of $\gamma$ determines which tokens are candidates for differentiated treatment. We adopt a quantile-based thresholding strategy: for each trajectory, we set $\gamma$ as the $q$-th quantile of the entropy values $\{H_t\}_{t=1}^T$, e.g., $q = 0.9$ selects the top 10\% highest-entropy tokens. This approach adapts to varying difficulty levels across problems and to entropy scale shifts across training stages; unlike a fixed absolute threshold, it remains active when the policy becomes more confident and absolute entropy values decrease.

\subsubsection{Trajectory Categorization and Advantage Shaping}
Given a set of sampled trajectories for each query, we first compute the trajectory-level advantage via GRPO normalization across the sampling group. We then classify trajectories into three categories based on correctness and length, applying differentiated advantage shaping to each:

\textbf{Category A: Correct and Concise Trajectories.} For trajectories that are both correct ($C(\tau) = 1$) and concise ($|\tau| = T_{\mathrm{ref}}$), we preserve the original advantage for all tokens to reinforce high-quality reasoning behavior:
\begin{equation}
\tilde{A}_{t} = A(\tau), \quad \forall t \in \{1, \ldots, |\tau|\},
\end{equation}

\textbf{Category B: Incorrect Trajectories.} For incorrect trajectories ($C(\tau) = 0$), we preserve the original advantage for all tokens to maintain learning pressure toward correctness, without adding AES-specific suffix penalties before the policy has found a valid solution:
\begin{equation}
\tilde{A}_{t} = A(\tau), \quad \forall t \in \{1, \ldots, |\tau|\},
\end{equation}

\textbf{Category C: Correct but Overthinking Trajectories.} For trajectories that are correct but exhibit overthinking ($C(\tau) = 1$ and $|\tau| > T_{\mathrm{ref}}$), we selectively penalize high-entropy tokens while sparing the rest. Specifically, we re-compute the token-level advantage as:
\begin{equation}
\tilde{A}_{i,t} = 
\begin{cases}
A_i - \beta_{\mathrm{AES}} \cdot \eta_i \cdot \lambda(t), 
& \text{if } t \in \mathcal{S}(\tau_i) \text{ and } t>T_{\mathrm{ref}}, \\
A_i, & \text{otherwise}.
\end{cases}
\end{equation}
where $\beta_{\mathrm{AES}}$ is the AES penalty coefficient, $\eta_i$ is a group-wise min-max normalized trajectory-level excess-length factor, $\lambda(t)$ is a suffix-position scaling factor, and $\mathcal{S}(\tau_i)$ denotes the set of high-entropy tokens; full definitions are provided in Appendix~\ref{app:scaling}.
This formulation ensures that tokens most likely to be associated with superfluous branching receive reduced advantage values, while those outside the overlength high-entropy suffix, including high-entropy positions before $T_{\mathrm{ref}}$, retain their original learning signals.

The reference length $T_{\mathrm{ref}}$ is defined as the length of the shortest correct trajectory within each sampling group. It serves only as the boundary for identifying overlength suffixes, rather than as a token-level imitation target. If the group contains no correct trajectory, AES disables selective shaping and retains the original GRPO advantages for all trajectories in the group.

\subsection{Entropy and Expected Reasoning Length}

We relate token-level entropy to future reasoning cost through a simple branching view of generated reasoning trajectories. The central intuition is that high-entropy tokens often occur at states where multiple continuations are plausible, and such states may lead the model into additional verification or backtracking.

\textbf{Effective branching factor.}
Let $s_t = (x, y_{<t})$ denote the state at generation step $t$, comprising the input and all previously generated tokens. The conditional entropy $H_t = H(\pi_\theta(\cdot \mid s_t))$ captures the model's uncertainty over the next token. We define the \emph{effective branching factor} as $B_t := \exp(H_t)$, which represents the effective number of plausible continuations available at state $s_t$.

\textbf{Branching model of overthinking.}
We model the reasoning process as a stochastic branching process. At each step $t$, the model either proceeds toward termination or initiates a verification/backtracking phase with probability $q_t$. We use the following local parameterization:
\begin{equation}
q_t = 1-\exp(-\alpha B_t),
\end{equation}
where $\alpha>0$ controls how strongly the effective branching factor increases the probability of additional reasoning. Under the branching assumptions detailed in Appendix~\ref{app:branching-proof}, this yields the following local relation:

\begin{proposition}\label{prop:entropy-length-main}
Let $L_t = \mathbb{E}[T-t \mid s_t]$ denote the expected remaining trajectory length at step $t$. Under the local homogeneous branching model with $q_t=1-\exp(-\alpha B_t)$, if $\alpha B_t \leq 1$, then there exist constants $c_1,c_2>0$ such that
\begin{equation}
c_1 B_t \leq L_t \leq c_2 B_t .
\end{equation}
Since $B_t=\exp(H_t)$, the expected residual length locally scales with the exponential of token-level entropy under this branching approximation.
\end{proposition}

This proposition provides an interpretation of why high-entropy tokens can serve as useful indicators of branching-prone regions in reasoning trajectories. Consistent with this view, longer correct trajectories exhibit higher mean and 90th-percentile token entropy than the shortest correct trajectory in the same sampling group (Appendix~\ref{app:entropy_diff}). AES uses this signal conservatively by applying credit reduction only to high-entropy tokens in the overlength suffix of correct trajectories, while preserving earlier high-entropy tokens that may be necessary for productive exploration.

\section{Experiments}

\begin{table*}[t]
\centering
\caption{Main results on mathematical reasoning benchmarks. We report accuracy (\%) and average response length (Len). In the Mean columns, \textbf{bold} and \underline{underlined} values indicate the best and second-best aggregate values among rows shown, respectively. }
\label{tab:main_results_math}

\resizebox{\textwidth}{!}{%
\begin{tabular}{llcccccccccc}
\toprule
\multirow{2}{*}{Model} & \multirow{2}{*}{Method} & \multicolumn{2}{c}{GSM8K} & \multicolumn{2}{c}{MATH-500} & \multicolumn{2}{c}{AMC} & \multicolumn{2}{c}{OlympiadBench} & \multicolumn{2}{c}{Mean} \\
\cmidrule(lr){3-4} \cmidrule(lr){5-6} \cmidrule(lr){7-8} \cmidrule(lr){9-10} \cmidrule(lr){11-12}
& & Acc & Len & Acc & Len & Acc & Len & Acc & Len & Acc & Len \\
\midrule
\multirow{7}{*}{\textit{R1-1.5B}} 
& Base Model & 82.1 & 1813 & 83.5 & 5736 & 49.3 & 6820 & 44.8 & 11497 & 64.9 & 6467 \\
& Efficient Reason & 78.1 & 324 & 81.8 & 2522 & 46.6 & 3345 & 45.2 & 6302 & 62.9 & \underline{3123} \\
& LC-R1 & 81.2 & 745 & 80.2 & 2505 & 50.3 & 3432 & 43.0 & 5836 & 63.7 & 3130 \\
& Laser & 87.0 & 1778 & 84.8 & 3321 & 51.4 & 4259 & 49.4 & 5819 & \underline{68.2} & 3794 \\
& ThinkPrune & 85.7 & 1131 & 83.7 & 2791 & 51.1 & 3527 & 46.3 & 5520 & 66.7 & 3242 \\
& Thinkless & 85.1 & 986 & 82.3 & 3586 & 51.8 & 4463 & 47.0 & 7359 & 66.6 & 4099 \\
\rowcolor{blue!6}
& \textbf{AES (Ours)} & 87.4 & 1121 & 85.7 & 2643 & 51.6 & 3274 & 49.9 & 5156 & \textbf{68.7} & \textbf{3048} \\
\midrule
\multirow{6}{*}{\textit{R1-7B}} 
& Base Model & 90.1 & 1032 & 92.7 & 3917 & 55.3 & 5100 & 58.1 & 8800 & \underline{74.1} & 4712 \\
& Efficient Reason & 87.8 & 278 & 90.7 & 2189 & 50.9 & 3050 & 55.9 & 6063 & 71.3 & 2895 \\
& LC-R1 & 87.3 & 423 & 89.1 & 1552 & 50.3 & 2446 & 55.0 & 4440 & 70.4 & \textbf{2215} \\
& GRPO-LEAD & 90.1 & 814 & 92.3 & 3220 & 55.4 & 3991 & 58.4 & 6767 & \underline{74.1} & 3698 \\
& VeriThinker & 92.5 & 763 & 92.1 & 2589 & 51.3 & 3589 & 58.2 & 6292 & 73.5 & 3308 \\
\rowcolor{blue!6}
& \textbf{AES (Ours)} & 92.9 & 734 & 91.7 & 2103 & 57.4 & 2897 & 58.9 & 4305 & \textbf{75.2} & \underline{2510} \\
\bottomrule
\end{tabular}%
}
\end{table*}

\begin{table*}[t]
\centering
\caption{Results on competition-level mathematics and out-of-domain benchmarks. We report accuracy (\%) and average response length (Len). In the Mean columns, \textbf{bold} and \underline{underlined} values indicate the best and second-best aggregate values among rows shown, respectively.}
\label{tab:main_results_hard}
\resizebox{\textwidth}{!}{%
\begin{tabular}{llcccccccccc}
\toprule
\multirow{2}{*}{Model} & \multirow{2}{*}{Method} & \multicolumn{2}{c}{AIME24} & \multicolumn{2}{c}{AIME25} & \multicolumn{2}{c}{LSAT} & \multicolumn{2}{c}{GPQA} & \multicolumn{2}{c}{Mean} \\
\cmidrule(lr){3-4} \cmidrule(lr){5-6} \cmidrule(lr){7-8} \cmidrule(lr){9-10} \cmidrule(lr){11-12}
& & Acc & Len & Acc & Len & Acc & Len & Acc & Len & Acc & Len \\
\midrule
\multirow{7}{*}{\textit{R1-1.5B}} 
& Base Model & 27.9 & 16645 & 23.1 & 16623 & 22.8 & 14549 & 34.1 & 10090 & 27.0 & 14477 \\
& Efficient Reason & 29.2 & 10517 & 22.1 & 9755 & 23.2 & 7694 & 33.0 & 6037 & 26.9 & 8501 \\
& LC-R1 & 24.6 & 8684 & 22.1 & 8526 & 24.4 & 11711 & 35.5 & 8575 & 26.7 & 9374 \\
& Laser & 34.4 & 8492 & 22.5 & 8368 & 24.1 & 8059 & 35.5 & 5842 & \underline{29.1} & 7690 \\
& ThinkPrune & 30.4 & 7690 & 22.5 & 7750 & 25.5 & 6969 & 34.6 & 5785 & 28.3 & \underline{7049} \\
& Thinkless & 29.2 & 10493 & 23.5 & 10491 & 25.7 & 11214 & 30.9 & 10358 & 27.3 & 10639 \\
\rowcolor{blue!6}
& \textbf{AES (Ours)} & 31.5 & 7401 & 24.1 & 7933 & 25.9 & 7047 & 38.0 & 5194 & \textbf{29.9} & \textbf{6894} \\
\midrule
\multirow{6}{*}{\textit{R1-7B}} 
& Base Model & 53.1 & 13107 & 37.3 & 14485 & 41.1 & 10854 & 48.9 & 8079 & 45.1 & 11631 \\
& Efficient Reason & 50.8 & 9855 & 35.6 & 11075 & 42.0 & 7836 & 48.3 & 5679 & 44.2 & 8611 \\
& LC-R1 & 52.5 & 6915 & 32.1 & 8193 & 43.3 & 6954 & 49.7 & 4756 & 44.4 & \underline{6705} \\
& GRPO-LEAD & 54.2 & 10437 & 36.9 & 11621 & 42.3 & 8622 & 50.2 & 6754 & \underline{45.9} & 9359 \\
& VeriThinker & 54.6 & 9927 & 37.9 & 10548 & 29.5 & 6073 & 46.8 & 4456 & 42.2 & 7751 \\
\rowcolor{blue!6}
& \textbf{AES (Ours)} & 54.9 & 7075 & 38.4 & 7209 & 43.9 & 5318 & 51.3 & 4157 & \textbf{47.1} & \textbf{5940} \\
\bottomrule
\end{tabular}
}
\end{table*}

\subsection{Main Results}

\textbf{Overall Performance.} AES consistently achieves the best trade-off between accuracy and efficiency across both model scales, while using the shortest mean length in three settings and the second-shortest mean length in the remaining setting. On mathematical reasoning benchmarks (\cref{tab:main_results_math}), AES attains mean accuracies of 68.7\% for R1-1.5B and 75.2\% for R1-7B, while achieving length reductions of 52.9\% and 46.7\% compared to the base models, respectively. On the more challenging competition-level and out-of-domain benchmarks (\cref{tab:main_results_hard}), AES improves mean accuracy by 2.9\% for R1-1.5B and 2.0\% for R1-7B, while reducing response length by 52.4\% and 48.9\%, respectively.
The tables report accuracy and length as separate axes rather than a formal Pareto-frontier analysis. We therefore interpret AES primarily as an average accuracy-efficiency improvement.

\textbf{Comparison with Uniform Penalty Methods.} Methods that apply uniform length penalties, such as Efficient Reason and LC-R1, achieve large length reductions but suffer from notable accuracy degradation. For instance, on R1-1.5B, Efficient Reason reduces length by 51.7\% on mathematical benchmarks but incurs a 2.0\% accuracy drop (64.9\% to 62.9\%). Similarly, LC-R1 exhibits a 1.2\% accuracy decrease despite moderate length savings. These results support the concern that uniform penalties can cause credit confounding by suppressing both essential reasoning steps and superfluous elaboration. 
By contrast, AES achieves comparable or greater length reductions while improving accuracy, indicating that entropy-guided selective penalization better preserves task-relevant reasoning while discouraging redundant generation.

\textbf{Comparison with Adaptive Methods.} Among adaptive approaches, GRPO-LEAD and VeriThinker maintain competitive accuracy but provide limited efficiency gains. GRPO-LEAD achieves competitive accuracy but reduces length by only 21.5\% compared to the base model. VeriThinker shows similar patterns, trading modest length reduction for accuracy preservation. Thinkless, which adaptively selects reasoning modes, achieves only modest length reduction on challenging benchmarks (10639 vs. 14477 for R1-1.5B). 
AES outperforms these methods by applying credit reduction to overlength high-entropy suffix tokens, achieving superior efficiency without sacrificing accuracy.

\textbf{Performance on Challenging Benchmarks.} The advantages of AES become more pronounced on difficult tasks. On AIME24 and AIME25, which represent competition-level mathematics, AES provides competitive accuracy while substantially shortening responses. For R1-7B, AES improves AIME25 accuracy from 37.3\% to 38.4\% while reducing length by 50.2\%. On out-of-domain benchmarks, AES shows positive transfer within the evaluated model family. On GPQA, AES improves accuracy by 3.9\% for R1-1.5B and 2.4\% for R1-7B, with length reductions exceeding 48\%. On LSAT, AES achieves 3.1\% accuracy improvement for R1-1.5B and 2.8\% for R1-7B, while reducing response length by 51.6\% and 51.0\%, respectively. These results suggest that AES improves the average accuracy-efficiency trade-off by reducing empirically redundant continuation.

\subsection{Analysis and Ablation Studies}

\begin{figure*}[t]
    \centering
    \begin{subfigure}[b]{0.32\textwidth}
        \includegraphics[width=\linewidth]{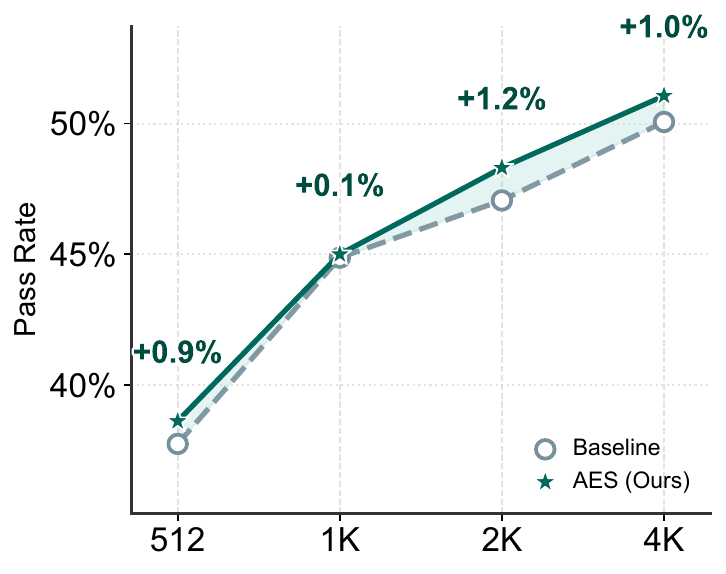}
        \caption{AMC}
    \end{subfigure}
    \hfill
    \begin{subfigure}[b]{0.32\textwidth}
        \includegraphics[width=\linewidth]{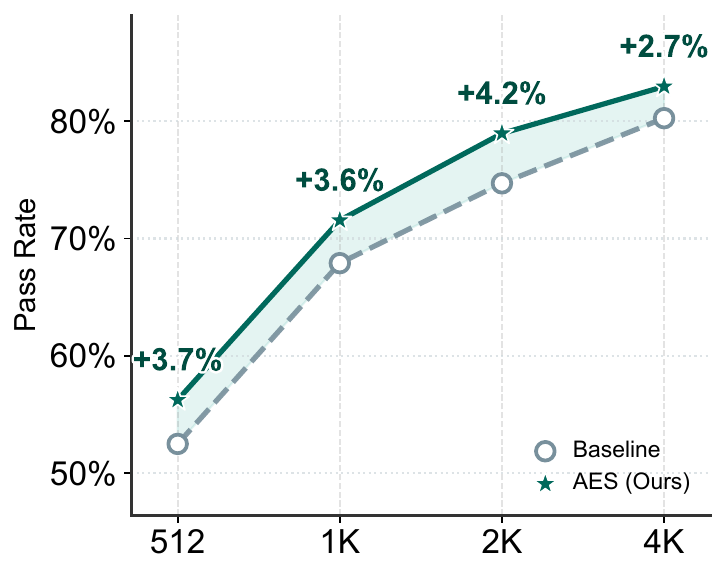}
        \caption{MATH-500}
    \end{subfigure}
    \hfill
    \begin{subfigure}[b]{0.32\textwidth}
        \includegraphics[width=\linewidth]{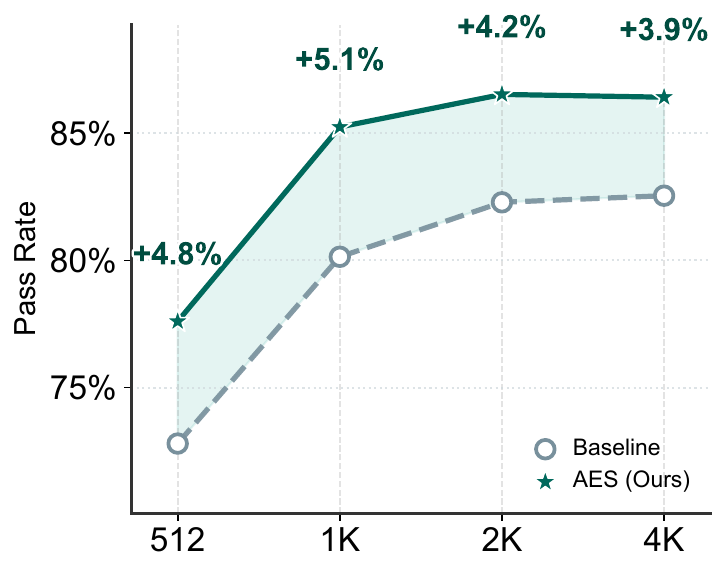}
        \caption{GSM8K}
    \end{subfigure}
    
    \caption{\textbf{Performance Comparison under Varying Token Budgets.} We compare the pass rates of AES and the baseline model under budget forcing, with maximum output lengths ranging from 512 to 4,096 tokens.}
    \label{fig:budget}
\end{figure*}

\textbf{Token Budget Constraints.} 
To examine whether AES produces more concise reasoning paths, we evaluate performance under varying token budgets (512--4K) against the budget forcing baseline~\citep{s1}. As shown in \cref{fig:budget}, AES has higher pass rates than budget forcing across the evaluated budget levels. On GSM8K, AES achieves 85.2\% at 1K tokens compared to 80.1\% for budget forcing (+5.1\%), with similar improvements observed on MATH-500 (+3.6\% at 1K) and AMC (+1.2\% at 2K). Notably, on GSM8K, AES at a 1K budget already exceeds budget forcing at 2K. These results suggest that AES not only improves token efficiency, but also learns to construct more streamlined reasoning trajectories, achieving competitive accuracy with substantially fewer tokens.

\begin{table}[t]
\centering
\caption{We evaluate the contribution of each design choice in AES on MATH-500. ``Random Selection'' penalizes randomly selected tokens with the same budget as the high-entropy set. ``w/o Scaling (Trajectory-Level)'' applies uniform penalty weights across all overthinking trajectories. ``w/o Scaling (Token-Level)'' uses identical penalty weights for all high-entropy tokens within each trajectory.}
\label{tab:ablation}
\small 
\begin{tabular}{lcc}
\toprule
Variant & Accuracy & Tokens \\
\midrule
Full AES & \textbf{91.7} & 2103 \\
\midrule
w/o Scaling (Trajectory-Level) & 91.1 & 2271 \\
w/o Scaling (Token-Level) & 90.0 & 2546 \\
Random Selection & 88.5 & 1734 \\
\bottomrule
\end{tabular}
\end{table}

\textbf{Component Analysis.}
\cref{tab:ablation} presents an ablation study on the key components of AES. Random selection yields the lowest accuracy (88.5\%), indicating that the entropy criterion is more informative than a matched random token set in this setting. This suggests that indiscriminate penalization can substantially reduce response length, but may also suppress reasoning steps that are necessary for problem solving, thereby degrading accuracy. Removing trajectory-level scaling reduces accuracy to 91.1\%, suggesting that differentiated penalty weights based on the degree of overthinking across trajectories contribute meaningfully to performance. Similarly, ablating token-level scaling results in 90.0\% accuracy with increased token usage, suggesting that adaptive penalties within each trajectory help the model focus on high-uncertainty suffix positions. The full AES achieves the best accuracy (91.7\%) with moderate token consumption, supporting the usefulness of both hierarchical scaling mechanisms.

\begin{table}[t]
\centering
\begingroup
\caption{Sensitivity to AES hyperparameters on R1-7B. We report mean accuracy (\%) and average response length on the mathematical benchmark group and the hard/OOD benchmark group.}
\label{tab:sensitivity}
\small 
\begin{tabular}{lcccc}
\toprule
Setting & Hard/OOD Acc & Hard/OOD Len & Math Acc & Math Len \\
\midrule
$\beta_{\mathrm{AES}}=1.0$ & \textbf{47.2} & 6544 & \textbf{75.2} & 3016 \\
$\beta_{\mathrm{AES}}=1.5$ & 46.9 & 6045 & 75.1 & 2667 \\
$\beta_{\mathrm{AES}}=2.0$ & 47.1 & 5940 & \textbf{75.2} & 2510 \\
$\beta_{\mathrm{AES}}=2.5$ & 46.8 & 5432 & 74.7 & 2273 \\
\midrule
$q=0.60$ & 46.1 & 5537 & 74.3 & 2061 \\
$q=0.75$ & 47.0 & 5770 & 74.9 & 2149 \\
$q=0.90$ & 47.1 & 5940 & \textbf{75.2} & 2510 \\
\bottomrule
\end{tabular}%
\endgroup
\end{table}

\textbf{Hyperparameter Sensitivity.}
\cref{tab:sensitivity} evaluates the sensitivity of AES to the penalty coefficient $\beta_{\mathrm{AES}}$ and the entropy threshold $q$. Overall, AES remains robust across a range of hyperparameter settings. Increasing $\beta_{\mathrm{AES}}$ generally leads to shorter responses, although overly large penalty weights can introduce a modest reduction in accuracy. Similarly, decreasing $q$ expands the set of penalized tokens and can further reduce generation length; however, the resulting accuracy trends are not strictly monotonic across benchmark groups, suggesting that the optimal degree of token suppression depends on task-specific reasoning characteristics. Based on this trade-off, we set $\beta_{\mathrm{AES}}=2.0$ and $q=0.9$ as a balanced configuration that achieves substantial length reduction while preserving strong accuracy. We keep the group size fixed at $G=8$, following standard GRPO practice. Since AES modifies token-level credit assignment after group-level rewards are computed, its mechanism is largely orthogonal to the choice of group size.

\section{Conclusion}
In this work, we introduce AES, a post-training method for mitigating overthinking in LRMs. We first conduct an empirical analysis of reasoning trajectories and observe that high-entropy tokens often function as transition points associated with unnecessary reasoning expansion. Motivated by this observation, AES selectively penalizes high-entropy suffix tokens that are likely to induce redundant elaboration, while preserving tokens that support productive reasoning. We further provide a branching interpretation that relates token-level entropy to expected residual reasoning cost under local modeling assumptions. Extensive experiments demonstrate that AES reduces response length by approximately 50\% while simultaneously improving accuracy, substantially outperforming existing efficiency-oriented methods.

\begin{ack}
This work was supported in part by the National Natural Science Foundation of China under Grant 62372132. The authors would also like to thank Huawei Ascend Cloud Ecological Development Project for providing high-performance Ascend 910 processors. The authors declare no competing interests.
\end{ack}

\bibliographystyle{plainnat}
\bibliography{ref}

\newpage
\appendix

\section{Scope and Limitations}
Our evidence is empirical and tied to a fixed evaluation protocol. The main results are most direct for DeepSeek-R1-Distill-Qwen 1.5B and 7B models, and cross-family validation on larger LRMs remains an important direction for future work. 

AES inherits a basic ambiguity shared by efficient-reasoning methods: additional reasoning can be either redundant verification or genuinely useful exploration. The shortest correct in-group response provides a simple within-prompt efficiency reference, but it is not guaranteed to be the only valid reference, especially for problems near the model's capability boundary. Finally, AES is designed to address uncertainty-driven branching behavior rather than low-entropy repetitive loops; the latter were rare in our rollouts, but remain outside the method's direct scope.

\section{Related Work}

\textbf{Large Reasoning Models.} Chain-of-Thought prompting~\citep{cot,cot_survey} has enabled LLMs to solve complex problems through step-by-step reasoning. Building on this foundation, recent slow-thinking models~\citep{openai_o1,deepseek,qwq} leverage reinforcement learning with verifiable rewards~\citep{rlvr} to generate extended reasoning trajectories. While effective, these models often suffer from \textit{overthinking}~\citep{overthinking}, producing verbose outputs with redundant verification steps that inflate computational costs.

\textbf{Efficient Reasoning.} To mitigate overthinking, prior work has explored several strategies to enhance reasoning efficiency~\citep{efficient_survey}. A prominent line of research employs reinforcement learning to explicitly penalize verbosity by incorporating length-based penalties into the reward function~\citep{o1pruner, l1, thinkprune, yeo2025demystifying, aalc}. However, these methods apply uniform penalties across all tokens, failing to distinguish essential reasoning steps from superfluous ones. Other approaches focus on inference-time prompting to guide more concise generation~\citep{chain_of_draft, sketch} or fine-tuning on compressed reasoning chains~\citep{c3ot, pause, icot}. Recent reasoning-efficiency work also explores early stopping and alternative execution structures: S-GRPO constructs serial early-exit groups and decaying rewards to encourage earlier correct answers~\citep{sgrpo2025}; Think-or-Not uses information-theoretic signals for entropy-based adaptive halting at inference time~\citep{think_or_not2025}; and SPRINT reorganizes reasoning trajectories into planning and parallel execution rounds to reduce sequential tokens~\citep{sprint2025}. These methods show that regulating intermediate reasoning is a central theme in efficient LRMs. Token entropy has also been used to locate decision points in reasoning, either to concentrate RLVR updates on high-entropy forking tokens~\citep{wang2025forking} or to prune low-surprisal steps for efficient chain-of-thought distillation~\citep{zeng2025surprisal}, whereas AES uses it conditionally to reduce credit only for high-entropy suffix tokens of overlong correct trajectories. Beyond shortening outputs, linear attention reduces the per-token cost of generation~\citep{meng2026polaformer, meng2025nalaformer} and has been used to model reasoning as state transitions~\citep{zhang2026statetransition}. AES occupies a complementary point in this design space: it keeps the standard rollout format, adds no test-time controller, and performs token-level advantage shaping only on high-entropy suffix positions in correct overlong samples.

\section{Details of Suffix-Aware Scaling}
\label{app:scaling}

For each correct overthinking trajectory $\tau_i$, we compute its relative excess length:
\begin{equation}
\Delta_i = \frac{|\tau_i| - T_{\mathrm{ref}}}{T_{\mathrm{ref}}}.
\end{equation}
Let $\mathcal{O}=\{i:C(\tau_i)=1,\ |\tau_i|>T_{\mathrm{ref}}\}$ denote the set of correct overthinking trajectories in the same sampling group. We apply min-max normalization within this set:
\begin{equation}
\eta_i =
\frac{\Delta_i - \min_{j\in\mathcal{O}}\Delta_j}
{\max_{j\in\mathcal{O}}\Delta_j - \min_{j\in\mathcal{O}}\Delta_j + \epsilon},
\end{equation}
where $\epsilon$ is a small constant for numerical stability. If all trajectories in $\mathcal{O}$ have the same excess length, we set $\eta_i=1$.

For token position $t$, we define the token-level suffix scaling factor as
\begin{equation}
\lambda(t) = \log\left(1 + \max\left(0, \frac{t - T_{\mathrm{ref}}}{T_{\mathrm{ref}}}\right)\right).
\end{equation}
The trajectory-level factor $\eta_i$ differentiates correct overthinking trajectories according to their relative excess length within the sampling group, while the token-level factor $\lambda(t)$ differentiates high-entropy suffix tokens according to their distance from the reference boundary. The logarithmic form makes $\lambda(t)$ grow sublinearly, preventing very late tokens in extremely long rollouts from dominating the update. Together, these two factors apply mild or no credit reduction near the efficient reference boundary, while assigning stronger but controlled penalties to later high-entropy suffix tokens in trajectories with larger excess length. This focuses shaping on transition points that are more likely to correspond to redundant verification or superfluous branching, consistent with the truncation results in \cref{efficiency_tradeoff}, while preserving pre-reference high-entropy tokens that may support necessary exploration.

\section{Experimental Settings} 
\textbf{Datasets.} For post-training, we utilize the DAPO-Math-17k dataset~\citep{dapo}. To comprehensively evaluate our method, we select several reasoning benchmarks spanning mathematical, logical, and scientific domains. For mathematical reasoning, we include GSM8K~\citep{gsm8k} for multi-step arithmetic word problems, and MATH-500~\citep{Math500,letverify} for competition-level mathematics. We further incorporate AMC~\citep{AMC23}, AIME24, and AIME25~\citep{AIME} from the American Mathematics Competitions, which demand creative problem-solving and non-routine strategies. OlympiadBench~\citep{olympiadbench} provides olympiad-level problems in mathematics and physics requiring rigorous multi-step proofs. For out-of-domain evaluation, we include LSAT~\citep{LSAT} for analytical deduction, and GPQA~\citep{GPQA} for graduate-level scientific reasoning in physics, chemistry, and biology.

\textbf{Comparison Models.} We compare AES against several baseline and published efficient-reasoning methods. 
The Base Model refers to the original DeepSeek-R1-Distill-Qwen-1.5B and -7B models~\citep{deepseek} (R1-1.5B and R1-7B) without any efficiency optimization.
Efficient Reason~\citep{efficient_reason} applies standard GRPO with a uniform length penalty term added to the reward. GRPO-LEAD~\citep{grpo_lead} introduces length-regularized rewards, explicit penalties for incorrect solutions, and difficulty-aware advantage reweighting to enhance both reasoning accuracy and efficiency. LC-R1~\citep{lcr1} combines length and compression rewards to encourage the model to retain only the most critical reasoning steps. ThinkPrune~\citep{thinkprune} trains the model under a limited token budget to enforce concise generation. Laser~\citep{laser} introduces a difficulty-aware length penalty reward that adapts penalization strength based on problem complexity. Thinkless~\citep{thinkless} adaptively selects between short-form and long-form reasoning depending on problem complexity. VeriThinker~\citep{verithinker} fine-tunes the model to verify the correctness of CoT solutions, thereby avoiding unnecessary self-reflection steps. Due to the availability of official implementations, we include different subsets of baselines for each model scale.

\textbf{Experimental Details.} Our framework is implemented using the open-source TRL library~\citep{trl}. We train with GRPO~\citep{reasoning_math} for 1,500 optimization steps using a learning rate of $1 \times 10^{-6}$, a batch size of 512, a group size of $G=8$, and a maximum completion length of 12,000 tokens. The GRPO clipping range is set to $[0.8, 1.28]$. We use separate coefficients for policy regularization and AES shaping: the GRPO KL coefficient is $\beta_{\mathrm{KL}}=0.001$, while the AES penalty coefficient is $\beta_{\mathrm{AES}}=2.0$ unless otherwise specified. The entropy quantile threshold is set to $q=0.9$, which selects the top 10\% highest-entropy tokens within each trajectory. We use public benchmark evaluation sets and benchmark-specific exact-match or rule-based answer checking after extracting the final answer from each response. All baselines are re-evaluated under the same decoding and evaluation protocol. All training experiments are conducted on 8 NVIDIA H100 GPUs.
For inference, we use temperature $0.6$, top-$p=0.95$. Each benchmark problem is evaluated with 16 independent sampled responses, without majority voting or best-of-$N$ selection. We report: (1) \textit{Accuracy}: average pass@1 percentage of problems solved correctly across all samples; (2) \textit{Token Count}: average number of tokens in the full generated response, including the reasoning trace and final answer. 

\textbf{Existing Assets and Licenses.} The third-party models, datasets, benchmarks, and software libraries used in this work are credited in the paper and used under their respective licenses or terms of use, including Apache-2.0, MIT, and CC-BY licenses where applicable. We do not redistribute third-party assets; users should obtain them from their official sources and comply with the corresponding license terms.

\section{Entropy of Longer versus Shortest Correct Trajectories}
\label{app:entropy_diff}

To check whether the entropy signal used by AES is associated with the trajectories it targets, we compare, within each activated sampling group, every longer correct trajectory with the shortest correct trajectory of the same group. 

\begin{table}[h]
\centering
\caption{Entropy comparison between longer correct trajectories and the shortest correct trajectory in the same activated sampling group (R1-7B, $G=16$). Each entry is the mean paired difference (longer $-$ shortest) of the per-trajectory mean token entropy or 90th-percentile token entropy.}
\label{tab:entropy_diff}
\small
\begin{tabular}{lcc}
\toprule
Dataset & Mean entropy diff. & 90th-pct. entropy diff. \\
\midrule
AIME25 & $+0.066$ & $+0.160$ \\
AIME24 & $+0.031$ & $+0.092$ \\
AMC & $+0.056$ & $+0.133$ \\
\bottomrule
\end{tabular}
\end{table}

\cref{tab:entropy_diff} reports the mean paired difference. On all three datasets the differences are positive for both the mean token entropy and the 90th-percentile token entropy, indicating that longer correct trajectories carry a heavier high-entropy tail than the efficient in-group reference.

\section{Derivation of the Entropy-Length Relationship}
\label{app:branching-proof}

We provide a derivation of Proposition~\ref{prop:entropy-length-main} under a  local branching model.

\paragraph{Model setup.}
At generation step $t$, let $B_t=\exp(H_t)$ denote the effective branching factor. We assume that the model initiates a verification/backtracking branch with probability
\begin{equation}
q_t = 1-\exp(-\alpha B_t),
\end{equation}
where $\alpha>0$. This parameterization is increasing in $B_t$ and saturates below one.

\paragraph{Residual length.}
Under a local homogeneous approximation, $q_t$ is treated as approximately constant over the immediate continuation. The number of verification/backtracking branches then follows a geometric process with expected count
\begin{equation}
\mathbb{E}[\text{\# branches}] = \frac{q_t}{1-q_t}.
\end{equation}
If $\ell_0>0$ denotes the baseline cost to terminate and $c>0$ denotes the expected token cost of each branch, the expected residual length is
\begin{equation}
L_t \approx \ell_0 + c\frac{q_t}{1-q_t}.
\label{eq:Lt-approx}
\end{equation}

\paragraph{Bounding the branching term.}
Substituting $q_t=1-\exp(-\alpha B_t)$ gives
\begin{equation}
\frac{q_t}{1-q_t}
=
\frac{1-e^{-\alpha B_t}}{e^{-\alpha B_t}}
=
e^{\alpha B_t}-1.
\end{equation}
Let $x=\alpha B_t$. In the moderate-entropy regime $x\leq 1$, we have
\begin{equation}
x \leq e^x-1 \leq 2x .
\end{equation}
Therefore,
\begin{equation}
\alpha B_t
\leq
\frac{q_t}{1-q_t}
\leq
2\alpha B_t.
\end{equation}
Substituting this into \eqref{eq:Lt-approx} yields
\begin{equation}
\ell_0+c\alpha B_t
\leq
L_t
\leq
\ell_0+2c\alpha B_t.
\end{equation}
Since $B_t=\exp(H_t)\geq 1$, the additive term can be absorbed into the upper constant:
\begin{equation}
L_t \leq (\ell_0+2c\alpha)B_t.
\end{equation}
The lower bound is already linear in $B_t$. Thus, for suitable constants $c_1,c_2>0$,
\begin{equation}
c_1B_t \leq L_t \leq c_2B_t,
\end{equation}
which gives $L_t=\Theta(B_t)=\Theta(\exp(H_t))$ in the local moderate-entropy regime.

\paragraph{Discussion of assumptions.}
The homogeneous approximation assumes that the branching probability remains approximately stable over the immediate continuation after state $s_t$. This approximation is most appropriate for a local analysis around a high-entropy state. It is reasonable when:
\begin{itemize}
    \item the model operates in a stable regime of uncertainty, so entropy does not fluctuate sharply across nearby continuation steps;
    \item the analysis focuses on the local effect of a high-entropy transition point.
\end{itemize}

This derivation is intended as an explanation of why high-entropy states can indicate larger future reasoning cost. More general time-varying branching processes may yield different functional forms, so the empirical truncation and ablation results remain the primary evidence for AES.


\newpage

\end{document}